\pdfoutput=1

\documentclass{article}
\usepackage{xcolor}
\usepackage{arxiv}

\usepackage{times}
\usepackage{soul}
\usepackage{url}
\usepackage[hidelinks]{hyperref}
\usepackage[utf8]{inputenc}
\usepackage[small]{caption}
\usepackage{graphicx}
\usepackage{amsmath}
\usepackage{amsthm}
\usepackage{booktabs}
\usepackage{algorithm}
\usepackage{algorithmic}
\usepackage[switch]{lineno}

\usepackage{subcaption}

\title{Exploring Spatial Language Grounding Through Referring Expressions}
\author{
Akshar Tumu$^1$
\and
Parisa Kordjamshidi$^2$\\
\affiliations
$^1$University of California San Diego, USA\\
$^2$Michigan State University, USA\\
\emails
\,
atumu@ucsd.edu,
kordjams@msu.edu
}

\begin{document}

\maketitle

\begin{abstract}
Spatial Reasoning is an important component of human cognition and is an area in which the latest Vision-language models (VLMs) show signs of difficulty. The current analysis works use image captioning tasks and visual question answering. In this work, we propose using the Referring Expression Comprehension task instead as a platform for the evaluation of spatial reasoning by VLMs. This platform provides the opportunity for a deeper analysis of spatial comprehension and grounding abilities when there is 1) ambiguity in object detection, 2) complex spatial expressions with a longer sentence structure and multiple spatial relations, and 3) expressions with negation (‘not’). In our analysis, we use task-specific architectures as well as large VLMs and highlight their strengths and weaknesses in dealing with these specific situations. While all these models face challenges with the task at hand, the relative behaviors depend on the underlying models and the specific categories of spatial semantics (topological, directional, proximal, etc.). Our results highlight these challenges and behaviors and provide insight into research gaps and future directions.
\end{abstract}
\section{Introduction}\label{sec:intro}
Vision-Language model (VLM) research has boomed in the recent past, owing to the enhanced user interaction and accessibility they provide. Models such as GPT 4o \footnote{\url{https://openai.com/index/hello-gpt-4o/}}, LLaVA~\cite{llava}, Google Gemini~\cite{gemini} have become adept at solving vision-language tasks such as Visual Question Answering (VQA), Image Captioning, and more. However, works like~\cite{vsr,subramaniamreclip,whatsup} show that VLMs still lack human-level ‘Spatial Reasoning' capabilities. Spatial reasoning involves comprehending relations that depict the absolute/relative position or orientation of an object, such as ‘left’, ‘above’, or ‘near’.

Most of the previous works confine their analysis to testing which models work well for spatial relations. We go further to analyze the comparative performance of these models for spatial categories that represent different orientational and positional relations between objects. A novel aspect of our work is the analysis of the effect of varying spatial composition (number of spatial relations) in the expressions on the performance of the models.

Previous works focused on spatial analysis with image captioning-related tasks, thus failing to locate the source of error in the presence of visual and linguistic ambiguity. To avoid this, we adopt the Referring Expression Comprehension (REC) task for our analysis. The REC models output bounding boxes around the target entity based on a natural language expression, the analysis of which could reveal the parts of the input that the models fail to comprehend.

For our analysis, we use the CopsRef dataset~\cite{copsref}, which is a complex dataset with visual ambiguity and multiple spatial relations in expressions. We focus our analysis on 51 spatial relations, categorized into 8 categories.

We test two popular VLMs - LLaVA~\cite{llava} and Grounding DINO~\cite{gdino}. We also included ‘MGA-Net’~\cite{mganet}, a model specifically designed for the REC task. The chosen models offer diversity in the evaluation as they differ in their architectural elements, training strategies, and input formats. We further compare these models with an object detector baseline to test if the images are truly complex and require elaborate referring expressions to ground the correct object.

Some of our important findings are as follows:

\noindent \textbf{(1)} Spatial relations contribute to more accurate grounding when added to other attributes of the objects in referring expressions. \noindent \textbf{(2)} Increasing the spatial complexity (no. of spatial relations) of an expression affects the performance of the VLMs, but models with explicit compositional learning components maintain the performance.
\noindent \textbf{(3)} Dynamic spatial relationships are difficult for all models to ground.
\noindent \textbf{(4)} The task-specific trained models find it easier to ground the geometric spatial relations such as left and right, while the VLMs perform better for ambiguous relations such as proximity. 
\noindent \textbf{(5)} All models struggle with handling negated spatial relations, but to varying degrees.

\begin{figure*}[!tb]
\centering
\begin{subfigure}{0.17\linewidth}
\includegraphics[width=\linewidth]{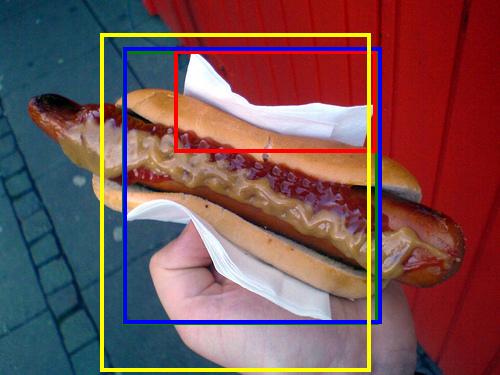}
\caption{The white napkin \\ that is wrapped around \\ the hot dog}
\label{2316628.jpg}
\end{subfigure} \hfill
\begin{subfigure}{0.17\linewidth}
\includegraphics[width=\linewidth, height=3cm]{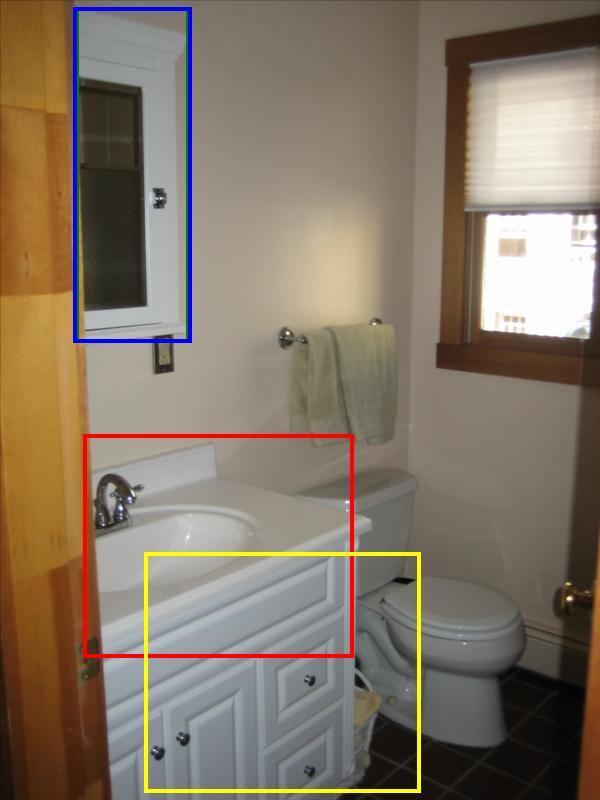}
\caption{The white box that is around the mirror}
\label{2064.jpg}
\end{subfigure} \hfill
\begin{subfigure}{0.17\linewidth}
\includegraphics[width=\linewidth]{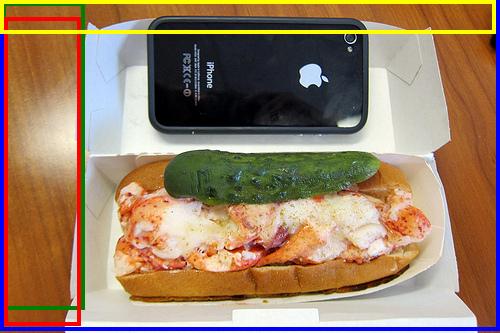}
\caption{The brown table that is to the left of the black cell phone}
\label{2413508.jpg}
\end{subfigure} \hfill
\begin{subfigure}{0.17\linewidth}
\includegraphics[width=\linewidth]{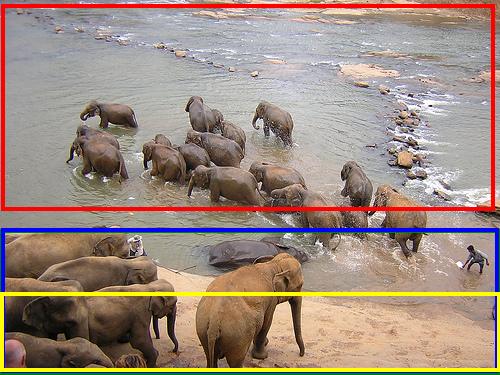}
\caption{The sandy shore that is near the murky water}
\label{2387165.jpg}
\end{subfigure} \hfill
\begin{subfigure}{0.17\linewidth}
\includegraphics[width=\linewidth]{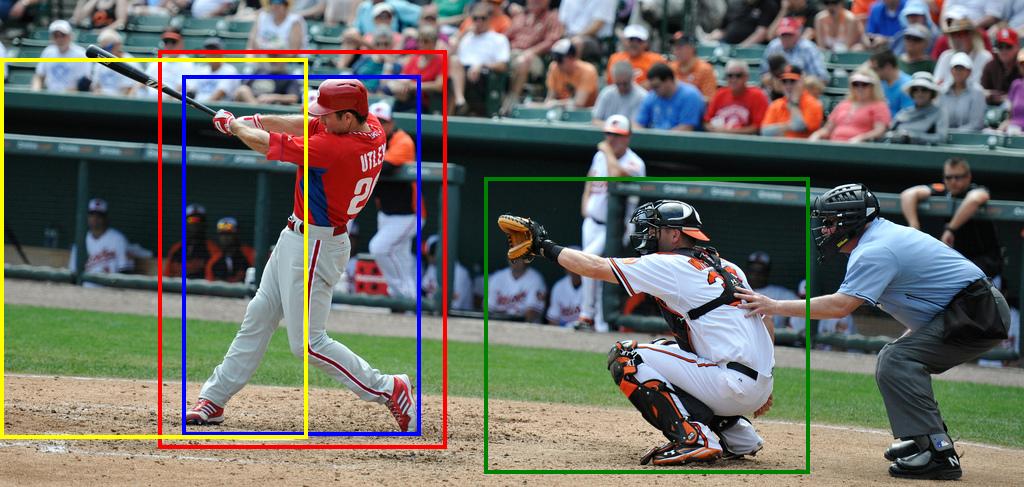}
\caption{The baseball player that is to the left of the black helmet and to the right of the home plate}
\label{1159745.jpg}
\end{subfigure} \hfill

\vspace{1em}
\begin{subfigure}{0.21\linewidth}
\includegraphics[width=\linewidth]{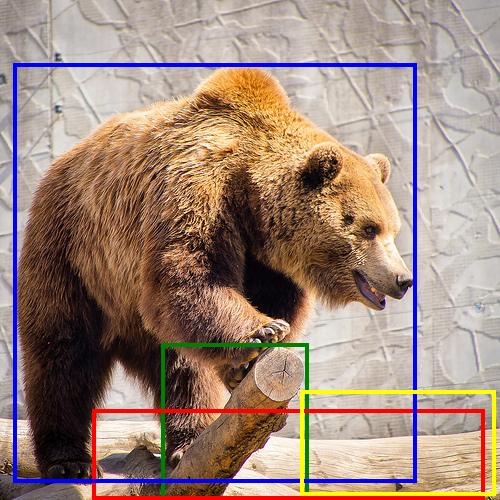}
\caption{The large branch that is to the right of the log that is behind the large bear}
\label{2324708.jpg}
\end{subfigure} \hfill
\begin{subfigure}{0.21\linewidth}
\includegraphics[width=\linewidth, height=3cm]{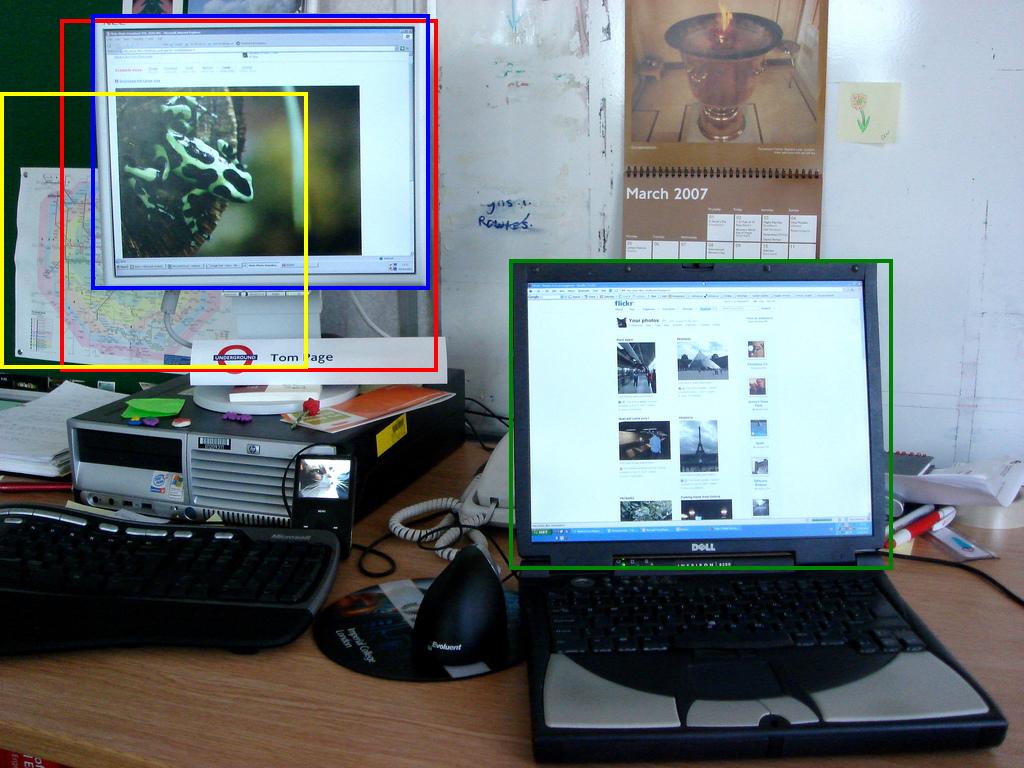}
\caption{The black monitor that is to the left of the keyboard or on the desk}
\label{285860.jpg}
\end{subfigure} \hfill
\begin{subfigure}{0.21\linewidth}
\includegraphics[width=\linewidth, height=3.5cm]{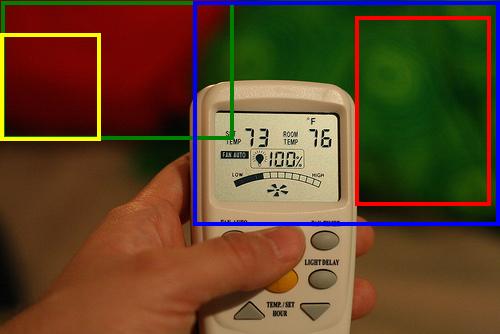}
\caption{The blanket that is not green and that is not on the bed}
\label{2363953.jpg}
\end{subfigure} \hfill
\begin{subfigure}{0.21\linewidth}
\includegraphics[width=\linewidth, height=3.5cm]{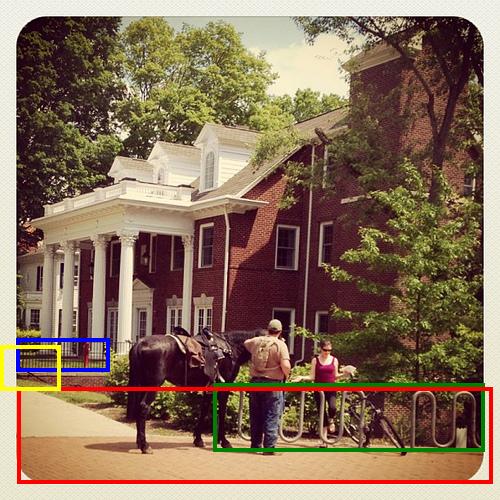}
\caption{The fence that is not black and that is not to the left of the man}
\label{2345193.jpg}
\end{subfigure}
\caption{Figures for Qualitative analysis. In the figures, the green box is the ground-truth bounding box. The red, blue, and yellow boxes are the output bounding boxes of MGA-Net, Grounding DINO, and LLaVA, respectively.}
\label{fig:example_figures}
\end{figure*}

\section{Related Work}\label{sec:rel work}
Previous works have performed a broad analysis of the ability of the VLMs to perform multimodal perception and reasoning tasks such as Spatial Reasoning, Multimodal conversation, etc. Works such as~\cite{cb300k,llava,mmbench,seedbench,mmvet,blink} introduce comprehensive real-world benchmarks to test multiple VLM capabilities.

~\cite{roschpi,subramaniamreclip} focus solely on spatial analysis of VLMs.~\cite{pic1k} go a step further to analyze the role of each modality in spatial reasoning. However, these works do not go deep to test the factors that affect the spatial reasoning ability of the VLMs. On the other hand,~\cite{vsr,film} perform a category-wise analysis of spatial relations. While the former categorize the relations based on their spatial properties, the latter categorize them as either simple, complex, implicit, or superlative.~\cite{sr2d} evaluate the models for the 4 spatial relations present in their dataset. Different from these works,~\cite{whatsup} analyze the effects of spatial biases in the datasets for REC task performance.~\cite{clipcdsm} analyze if the models are erring in recognizing objects, relations, or both. Some works like~\cite{cohn,liuother,mirzaee} focus on probing pre-trained LLMs or VLMs with text-only questions for spatial analysis. However, they do not test spatial grounding in the visual modality, a crucial aspect of our work.

Other closely aligned work includes Embodied Spatial Analysis which focuses on the effects of different perspectives and non-verbal cues on the spatial reasoning capabilities of VLMs~\cite{islam1,islam2}.
\noindent \paragraph{Task Complexity and Interpretability.} The works mentioned previously use image-caption agreement as their evaluation task. Due to the inherent limitations of this task, these works simplified the expressions to have only 2 objects and 1 spatial relation. Some works like~\cite{clipcdsm,subramaniamreclip,film} use synthetic datasets instead of real-world images to improve the interpretability of model output. But it simplifies the problem due to bounded expressivity (limited number of objects, attributes, and spatial relations). In our case, REC models output bounding boxes around the target objects. Analyzing the position and characteristics of the output object helps identify the parts of the input that the models fail to process. This enables comparative analysis of expressions with 0, 1, or more spatial relations, a unique feature of our work. The REC task also enables us to test the models over images of different visual complexities (single or multiple instances of objects in an image).
\begin{table*}[!ht]
\centering
\begin{tabular}{lrrrr}
\toprule
Dataset & Object Categories & Average length of & Average no. of objects & No. of spatial relations\\
 & & expression & per image & \\
\midrule
RefCOCO & 80 & 3.6 & 10.6 & 59 \\
RefCOCOg & 80 & 8.4 & 8.2 & 72 \\
CLEVR-Ref+ & 3 & 22.4 & 6.5 & 4 \\
CopsRef & 508 & 14.4 & 17.4 & 51 \\
\bottomrule
\end{tabular}
\caption{Statistics of Popular Referring Expression Comprehension datasets. For the last column, the relation types are taken from various resources explained in Section~\protect\ref{sec:rel work} and~\protect\cite{marchi}, in addition to the relations in Table~\protect\ref{table: our work cat}.}
\label{table: data-stats}
\end{table*}
\begin{table*}[!ht]
\centering
\begin{tabular}{lrl}
\toprule
Category & Number & Spatial Relations \\
\midrule
Absolute & 56 & on the right, on the left, in the middle, in the center, from the right, from the left\\
Adjacency & 14 & attached, against, on the side, on the back, on the front, on the edge \\
Directional & 29 & falling off, along, through, across, down, up, hanging from, coming from, around\\
Orientation & 0 & facing \\
Projective & 2361 & on top of, beneath, beside, behind, to the left, to the right, under, above, in front of,\\ & &  over, below, underneath \\ 
Proximity & 217 & by, close to, near \\ 
Topological & 1054 & connected, contain, with, surrounding, surrounded by, inside, between, touching,\\ & & out of, at, in, on \\
Unallocated & 56 & next to, enclosing \\
\bottomrule
\end{tabular}
\caption{Category-wise relation split and number of referring expressions in the CopsRef test set with 1 spatial relation in each category
}
\label{table: our work cat}
\end{table*}
\section{Dataset}\label{sec:dataset}
Table~\ref{table: data-stats} shows the key characteristics of some popular REC datasets. We chose CopsRef over RefCOCO and RefCOCOg due to the longer expression length and higher number of objects per image. Although CLEVR-Ref+ also provides a complex dataset, it is a synthetic dataset with limited expressiveness. Unlike other datasets, CopsRef's expressions go beyond describing the simple distinctive properties of the objects. CopsRef is also a highly spatial dataset as 90\% of expressions consist of spatial relations. Examples of such referring expressions and the corresponding images are given in Figure~\ref{fig:example_figures}.

Table~\ref{table: our work cat} shows the category-wise split of the 51 spatial relations we identified in the CopsRef test dataset. We utilize the categories introduced by~\cite{marchi} and replace the category 'Cardinal Direction' with 'Absolute'. For explanations of each category, refer to \cite{marchi}'s work and Appendix. Note that while there are no referring expressions with only one relation from the orientation category, these relations co-occur with relations from other categories in some expressions.

\begin{table}[!ht]
\centering
  \begin{tabular}{lrr}
    \toprule
    Category & No. of relations & No. of expressions\\
    \midrule
    None & 0 & 1202 \\
    One & 1 & 3787 \\
    Two-chained & 2 & 1324 \\
    Two-and & 2 & 3890 \\
    Two-or & 2 & 2203 \\
    Three & 3 & 180 \\
    \bottomrule
  \end{tabular}
  \caption{Frequency of occurrence of relations}
  \label{table: Cat-wise}
\end{table}

Table~\ref{table: Cat-wise} shows the number of expressions having 0, 1, 2, and 3 spatial relations. For expressions with 2 spatial relations, we have introduced three categories. The first category, ‘Two-chained’, consists of expressions in which spatial clauses are chained one after the other. The second category, ‘Two-and’, consists of expressions with two spatial clauses such that the referred object satisfies both of them. Finally, the ‘Two-or’ category consists of expressions with two spatial clauses such that the object referred to satisfies at least one of them. The referring expressions of Figures \ref{2324708.jpg}, \ref{1159745.jpg}, and \ref{285860.jpg} are examples of the three categories respectively.
\section{Approach}
In our analysis, we seek to answer the following research questions:\\
\textbf{RQ1.} Which spatial relation categories are more difficult to ground? \textbf{RQ2.} Do model characteristics/architecture influence the ease of grounding certain spatial relation categories over others? \textbf{RQ3.} Does using spatial relations improve grounding or make it more difficult? \textbf{RQ4.} How does the number of spatial relations in the expressions affect grounding in different types of models? \textbf{RQ5.} Are the REC models well-equipped to handle negated spatial relations?

To answer these questions, we explain our research methodology and the designed experiments in this section.
\subsection{Models Description}\label{subsec: res met - models}
We select three distinct models for our analysis such that they differ in key components like architecture, pre-training tasks, and input formats.
\noindent\paragraph{MGA-Net.}~\cite{mganet} is an REC task-specific model whose compositional learning architecture was designed to handle complex expressions. It decomposes a query using the soft attention mechanism and processes visual and linguistic information using dedicated modules to construct a relational graph among objects. Then, it uses a Gated Graph Neural Network to perform multi-step reasoning over the referring expression. We first implement the Faster-RCNN model~\cite{frcnn} to procure object proposals. Then, we generate the vector representations for these object proposals using a pre-trained ResNet-101 model. Considering the available computing resources, we omit the fourth (topmost) layer of the ResNet101 model to obtain a Partial CNN backbone. Finally, we train the model for ten epochs. We limit our training to ten epochs due to computational constraints.
\noindent\paragraph{Grounding DINO.}~\cite{gdino} It is an open-set object detector VLM with language support. It has a vision and a language backbone whose outputs are fused at multiple levels. Its contrastive loss for grounded pre-training makes it suitable for the REC task. We use the Swin-B vision backbone and the CLIP-text encoder for the language backbone.
\noindent\paragraph{LLaVA.}~\cite{llava} It is a general-purpose VLM that connects an open-set vision encoder from CLIP~\cite{clip} with a language decoder. The model is trained end-to-end which involves visual instruction tuning for aligning the vision and language modalities. We test LLaVA with a \textbf{Short prompt}: (USER: \textless image\textgreater\textbackslash n Give the bounding box for: "{Referring Expression}"\textbackslash nASSISTANT:) and a \textbf{Long prompt}: (USER: \textless image\textgreater\textbackslash n Provide the bounding box coordinates for the object described by the referring expression: "{Referring Expression}"\textbackslash n ASSISTANT:). Both prompts have a similar structure, but the second prompt is longer.
\noindent\paragraph{OWL-ViT}~\cite{owlvit} is an object detector baseline that only takes the target object's label as the input instead of the entire referring expression. It is an open-set object detector, which is required because CopsRef expressions involve entities from the Visual Genome~\cite{visualgenome} Scene Graphs, which have entities absent in common datasets used to train famous closed-set detectors like YOLO~\cite{yolo}. It also has a simple architecture with a Vision transformer and CLIP for aligning images and labels in a zero-shot manner, making it an ideal baseline.
\noindent\paragraph{Model Differences.} A key difference in the three main models can be seen in their input format. While Grounding DINO and LLaVA take the entire image as the input and perform bounding box regression to get object proposals, MGA-Net directly takes the externally detected bounding boxes as the input. Grounding DINO and LLaVA also have similarities in their architectures, as they both have vision and language backbones that are fed the entire image and text inputs. This is unlike MGA-Net, which has dedicated transformer architecture modules for visual, linguistic, and relative location components. However, Grounding DINO and MGA-Net show similarities in having grounded training/pre-training tasks while LLaVA only has general multimodal pre-training.
\subsection{Experimental Setting and Evaluation}\label{subsubsec: setting}
We create the following dataset test splits for evaluation and answering the earlier mentioned research questions, RQ1-RQ5. 
\subsubsection{Fine-grained Spatial Relations Split}\label{subsec: res met - spat rel cat}
In the test dataset, we split the expressions with 1 spatial relation using the categories shown in Table~\ref{table: our work cat}. Using the categories from Table~\ref{table: Cat-wise}, we split the remaining expressions based on the number of spatial relations they contain. Then, we rank the models based on their accuracy for each category.

To compare the models' performances across the categories, we employ a statistical test known as the Kendall Tau Independence Test. It evaluates the degree of similarity between two sets of ranks given to the same set of objects. We calculate the Kendall rank coefficient (\begin{math}\tau\end{math}) which yields the correlation between two ranked lists. Given \begin{math}\tau\end{math} value, we calculate the $z$ statistic, which follows standard normal distribution, as:
\begin{align}
    z = 3*\tau*\sqrt{n(n-1)}/\sqrt{2(2n+5)}.
\end{align}
Using the 2-tailed p-test at 0.05 level of significance, we test the following: \textbf{Null hypothesis}: There is no correlation between the two ranked lists. \textbf{Alternative hypothesis}: There is a correlation between the two ranked lists.
\subsubsection{Visual Complexity Split}\label{subsec: res met - vis_comp}
To observe the effect of visual complexity on model performance, we split the test dataset into two parts. The first part has images that have multiple instances of one or more objects mentioned in the associated referring expressions. The second part has images with at most one instance of every object mentioned in the expression. We perform this splitting by first collecting the entities in each expression using spaCy\footnote{\url{https://spacy.io/}} and then employing Grounding DINO to find the number of instances in the image for each of the collected entities.
\subsubsection{Negation Analysis Split}\label{subsec: res met - special cases}
In our analysis, we found that models have difficulties in grounding spatial expressions with negations. Therefore, we created a test split for a more accurate evaluation and a deeper analysis of negated spatial expressions. 
We collected expressions that include the keyword ‘not’ and divided them into two sets according to the number of occurring negations (1 or 2). Then, we collected those expressions for which all three models give an IoU of less than 0.5. For each expression, we perform a qualitative analysis to verify whether the errors are due to misinterpreting the negations or conflation of other errors. We limit our analysis to the results from the first run of the three models to facilitate the instance-wise analysis.
\section{Results}
\begin{table}[!ht]
\centering
  \begin{tabular}{lr}
    \toprule
    Model & Accuracy (\%) \\
    \midrule
    MGA-Net (Partial CNN) & $62.92 \pm 0.11$ \\
    Grounding DINO & $70.93 \pm 0.01$ \\
    LLaVA - Short Prompt & $34.96 \pm 0.03$ \\
    MGA-Net (Full CNN) & $61.22 \pm 0.15$\\
    LLaVA - Long Prompt & $33.79 \pm 0.01$ \\
    OWL-ViT & $56.34 \pm 0$\\
    \bottomrule
  \end{tabular}
\caption{Comprehension Accuracies}
\label{table: Comp-acc}
\end{table}
\begin{table*}[!htb]
\centering
\begin{tabular}{lrrrrrr}
\toprule
Category & MGA-Net (Acc \%) & Rank & Grounding DINO (Acc \%) & Rank & LLaVA (Acc \%) & Rank \\
\midrule
Absolute & $70.24 \pm 2.22$ & 1 & $82.14 \pm 0$ & 2 & $44.64 \pm 0$ & 4 \\
Adjacency & $52.38 \pm 3.37$ & 12 & $78.57 \pm 0$ & 4 & $50 \pm 0$ & 1 \\
Directional & $52.87 \pm 3.25$ & 11 & $65.52 \pm 0$ & 12 & $27.59 \pm 0$& 12 \\
Projective & $64.07 \pm 0.08$ & 4 & $69.12 \pm 0$ & 8 & $36.19 \pm 0.08$ & 6 \\
Proximity & $62.83 \pm 0.22$ & 8 & $80.65 \pm 0$ & 3 & $46.84 \pm 0.22$ & 3 \\
Topological & $67.32 \pm 0.49$ & 2 & $83.02 \pm 0$ & 1 & $48.51 \pm 0.09$ & 2 \\
Unallocated & $63.09 \pm 0.84$ & 6 & $75 \pm 0$ & 5 & $35.71 \pm 0$ & 7 \\
None & $62.39 \pm 0.58$ & 9 & $73.88 \pm 0$ & 6 & $42.89 \pm 0.17$ & 5 \\
Two-chained & $63.21 \pm 0.22$ & 5 & $70.67 \pm 0.03$ & 7 & $30.82 \pm 0$ & 9 \\
Two-and & $62.98 \pm 0.09$ & 7 & $68.45 \pm 0.01$ & 10 & $31.57 \pm 0.07$ & 8 \\
Two-or & $59.39 \pm 0.21$ & 10 & $68.97 \pm 0.01$ & 9 & $30.26 \pm 0.04$ & 10 \\
Three & $65.18 \pm 2.1$ & 3 & $67.78 \pm 0$ & 11 & $30.19 \pm 0.26$ & 11 \\
\bottomrule
\end{tabular}
\caption{Category-wise accuracy and ranking}
\label{table: Cat-wise res}
\end{table*}
\noindent\paragraph{Hardware.} For Grounding DINO, LLaVA, and the OWL-ViT baseline, we use the T4 GPU provided by Google Colaboratory for inference.\footnote{\url{https://colab.research.google.com/}} For MGA-Net, we use the NVIDIA GeForce GTX 1650 GPU for training. We run each model three times (both training and testing for MGA-Net, and inference for the VLMs and the baseline) to ensure the statistical significance of our results.
\noindent\paragraph{Evaluation Metrics.} We evaluate the models using the Intersection over Union (IoU) metric. Following previous works~\cite{mattnet,cmatterase}, we consider the output as a correct comprehension if the IoU is greater than 0.5. We calculate the \textit{comprehension accuracy} (referred to as accuracy) as the fraction of data points that have an IoU \textgreater 0.5.
\subsection{Evaluation on Referring Expressions}
From Table~\ref{table: Comp-acc}, we can observe that Grounding DINO and MGA-Net outperform the OWL-ViT baseline, with the former achieving the highest accuracy in grounding the referring expressions. However, we also tried training MGA-Net with the full ResNet-101 visual backbone (Full CNN) instead of the partial backbone (Partial CNN). We could only train this model for four epochs due to computational constraints. However, the model crossed 60\% test accuracy in just four epochs and was monotonically increasing. This shows that MGA-Net could potentially provide a better performance using adequate computational resources. To avoid unfair comparisons due to the training discrepancies, we focus our results on the relative performances of each model across different spatial relation categories rather than comparing the absolute performances.

For LLaVA, we used the prompts explained in Section~\ref{subsec: res met - models}. The shorter prompt gave a slightly better accuracy than the longer prompt. Hence, we used the short version for further experiments. The accuracy of LLaVA is less than both the other models and the baseline. Possible reasons are the lack of both bounding box regression and visual grounding instructions during pre-training.

Since we trained/tested each model for three runs, we report the average accuracy of the three runs and the standard deviation in the table. Since we re-train MGA-Net for each of these runs, there is a noticeable difference in model predictions in each run, leading to a slightly high standard deviation. However, we test the VLMs and the baseline zero-shot, leading to zero or near-zero standard deviation in the accuracies. This also follows for the future result tables.
\subsection{Evaluation on Fine-grained Relations}\label{subsec: Eval fgsp}
Table~\ref{table: Cat-wise res} shows a few general trends in results. The top 3-4 categories that each model performs the best for are categories with a single spatial relation. Among those, all 3 models perform well for the Topological and Absolute categories.

To answer \textbf{RQ1}, we observed that all the models struggle with the Directional relations. A possible reason is that the spatial configurations of the involved objects vary from image to image for the same spatial relation. This makes it difficult for the models to learn common patterns for recognizing these relations, resulting in low accuracy.

\begin{table}[!htb]
\centering
  \begin{tabular}{llrr}
    \toprule
        Model 1 & Model 2 & Coeff, & 2-tailed test \\
     & & Z-score & (Correlated) \\
    \midrule
    MGA-Net & GDINO & 0.18, 0.51 & No \\
    MGA-Net & LLaVA & 0.09, 0.26 & No \\
    GDINO & LLaVA & 0.73, 2.05 & Yes \\
    \bottomrule
    \end{tabular}
\caption{Kendall Tau Independence Test results for category-wise ranks. Coeff: Kendall Rank Coefficient, GDINO: Grounding DINO.}
\label{table: Kendall tau cat}
\end{table}

\begin{table*}[!htb]
\centering
  \begin{tabular}{lrrrrr}
    \toprule
    No. of relations & MGA-Net (Acc \%) & Grounding DINO (Acc \%) & LLaVA (Acc \%) & OWL-ViT Baseline (Acc \%)\\
    \midrule
    None & $62.39 \pm 0.59$ & $73.88 \pm 0$ & $42.89 \pm 0.17$ & $67.8 \pm 0$\\
    One & $64.85 \pm 0.13$ & $73.94 \pm 0$ & $40.33 \pm 0.01$ & $60.5 \pm 0$\\
    Two & $61.96 \pm 0.07$ & $69 \pm 0$ & $31.05 \pm 0.06$ & $52.39 \pm 0$\\
    Three & $65.18 \pm 2.1$ & $67.78 \pm 0$ & $30.19 \pm 0.26$ & $55 \pm 0$\\
    \bottomrule
    \end{tabular}
\caption{Relation frequency results and ranking}
\label{table: Freq-wise}
\end{table*}

\subsection{Impact of Multiple Spatial Relations}\label{subsec: results - model-wise-rel}
Table~\ref{table: Kendall tau cat} shows the Kendall Tau Independence test results for the three pairs of VLMs. We can observe that while the category-wise ranks of the VLMs (Grounding DINO and LLaVA) are correlated, MGA-Net's ranks aren't correlated with them. This motivates us to study the possible reasons behind the difference in the category-wise performances of MGA-Net and the VLMs. 

Among spatial categories of MGA-Net and VLMs, the major difference occurs with the Proximity and Projective categories. To answer \textbf{RQ2}, we can observe that the ‘Proximity’ category ranks third for both the VLMs but 8th for MGA-Net. On the other hand, ‘Projective’ has a higher rank for MGA-Net than both VLMs. We can see that MGA-Net prefers geometric spatial relations like left of, on top of, etc. as it takes the relative locations of bounding boxes as input which helps represent such relations. On the other hand, the two VLMs outperform in ambiguous relations which do not specify a clear distance or geometric direction, such as by/close-to. This is because the vision backbones of the VLMs utilize the entire image and help capture relations between a region in the image and its surrounding regions, unlike MGA-Net which only receives the detected bounding boxes as input.

To study further differences between MGA-Net and the VLMs, we design Table~\ref{table: Freq-wise} which shows the performance of the three models and the OWL-ViT baseline for expressions having different numbers of spatial relations. We observe that VLMs perform considerably better for expressions with 0/1 spatial relations compared to expressions with 2/3 spatial relations. This proves that VLMs find it difficult to ground multiple spatial relations. However, MGA-Net takes advantage of its compositional learning architecture to handle multi-step reasoning, resulting in a similar performance for all categories.

An interesting observation is that the performance of the baseline considerably drops for the ‘Two' and ‘Three' categories, even though the spatial relations aren't being passed as input to the baseline. The reason might be that 41.4\% of these images have multiple instances of objects, the impact of which is explained in the next section.

From~\ref{table: Freq-wise}, we can also compare the performance of the models for expressions with none and one spatial relation. We observe that LLaVA performs better for the former and MGA-Net for the latter. Grounding DINO gives a similar performance for both.

Now, to answer \textbf{RQ3}, we observe in Table~\ref{table: Cat-wise res} that among the seven categories of single spatial relations, MGA-Net and Grounding DINO perform better for five of those compared to expressions with no spatial relations. LLaVA also performs better for four such categories. Thus, we can conclude that in a setup involving visual and linguistic ambiguity (such as ours), spatial relations along with visual attributes often aid the models in grounding the expressions, compared to the attributes alone. This is also reinforced by the results of the baseline. From Table~\ref{table: Freq-wise}, we can observe that while the baseline gives the second-best performance for expressions with no spatial relations, it drops to the third place for expressions with one spatial relation with a 7.3\% reduction in performance. This is because the baseline doesn't have access to the spatial relations.

Finally, Table~\ref{table: Freq-wise} helps us answer \textbf{RQ4} as it shows the effect of increasing spatial relations on the performance of MGA-Net versus the VLMs (as discussed before).

\subsection{Impact of Visual Complexity}
\begin{table}[!ht]
\centering
\begin{tabular}{@{}l*{3}{l}@{}}
\toprule
Model    & Accuracy Single(\%) & Accuracy Multi(\%)\\ \midrule
MGA-Net   & $64.91 \pm 0.15$           & $59.61 \pm 0.04$   \\
G-DINO    & $72.54 \pm 0.01$           & $68.94 \pm 0.01$   \\
LLaVA     & $37.69 \pm 0.01$           & $30.43 \pm 0.1$    \\
OWL-ViT  & $59.71 \pm 0$              & $51.3 \pm 0$       \\
\bottomrule
\end{tabular}
\caption{Results for accuracy in different visual complexity settings.}
\label{tab:multi}
\end{table}
Out of 12586 test data points, we found that in the images of 4730 data points, there are multiple instances of objects mentioned in the referring expressions. Table~\ref{tab:multi} shows the accuracies of the three models and the OWL-ViT baseline for images with a single instance (‘Accuracy Single' column) and multiple instances (‘Accuracy Multi' column). The models perform better for the single instance images by 5.4\% on average compared to the multi-instance images. The 8.4\% performance drop of the baseline for multi-instance images proves that the images are indeed complex and require more than just the label as the input for grounding the right object. However, the 7.3\% performance drop of LLaVA, as compared to MGA-Net and Grounding DINO, shows that grounded pre-training also plays a crucial for multi-instance images.
\subsection{Impact of Negation}
\begin{table}[!ht]
\centering 
\begin{tabular}{@{}l*{2}{l}@{}}
\toprule
                 & Two Negations          & One Negation \\ \midrule
Total            & 73                     & 36                    \\
Total failure    & 59                     & 24                    \\
Grounding DINO   & 58                     & 23                    \\
LLaVA            & 29                     & 17                    \\
MGA-Net          & 35                     & 20                    \\ \bottomrule
\end{tabular}
\caption{Results for negations in expressions.}
\label{tab:negations}
\end{table}
We obtained 36 expressions with 1 ‘not’ and 73 expressions with 2 ‘not’s for which all models gave incorrect predictions. Table \ref{tab:negations} shows the total number of expressions we obtained with 1 and 2 negations. The ‘Total failure' row gives the number of instances for which models failed to recognize at least 1 negation. We can observe that Grounding DINO has the highest number of failure instances. LLaVA handles the negations better possibly due to the Vicuna \cite{vicuna} language backbone as it has a better language understanding (including negations) compared to Grounding DINO's CLIP text encoder. MGA-Net's training involves expressions with negations, making it more adept at recognizing them during testing. Hence, to answer \textbf{RQ5}, we observe that while all REC models face issues with recognizing negations, some models are comparatively better at handling them.

\begin{table}[!ht]
\centering
  \begin{tabular}{@{}l*{4}{l}|@{}}
    \toprule
    Models & Negations & Precision (\%) & Recall (\%) \\
    \midrule
    MGA-Net & 1 & 53.60 & 70.8 \\
    MGA-Net & 2 & 41.38 & 51 \\
    LLaVA & 1 & 64.54 & 47.23 \\
    LLaVA & 2 & 60.35 & 41 \\
    \bottomrule
  \end{tabular}
  \caption{Negation Metrics: MGA-Net vs. LLaVA}
  \label{table: Prec-Rec MGL}
\end{table}

Another interesting observation was for the outputs of MGA-Net and LLaVA models when they came close to the target object. From Table~\ref{table: Prec-Rec MGL}, we can see that while LLaVA has a better precision, MGA-Net has a better recall.
\section{Qualitative Analysis}
Here, we provide a qualitative analysis of certain issues faced by the models in handling referring expressions.
\subsection{Directional Relations}\label{subsec: qual analysis - Directional}
The expressions pertaining to Figures~\ref{2316628.jpg} and \ref{2064.jpg} consist of the same spatial relation (‘around’). In the first figure, the wrapping of the napkin around the hotdog only makes the napkin partially visible. But in the second figure, the white box around the mirror is almost entirely visible. This shows how the interpretation of ‘around’ is highly dependent on the configuration of the involved objects. For the first image, LLaVA fails to precisely localize the object, while MGA-Net only returns a part of the napkin that is visible. In the second image, both models fail to localize the object.
\subsection{Projective and Proximity Relations}\label{subsec: qual analysis - P&P}
Figure~\ref{2413508.jpg} shows an example of Projective relations (‘to the left’). MGA-Net succeeds in returning the correct part of the table that is to the left of the phone. While Grounding DINO simply returns the entire table, LLaVA identifies the wrong part. This shows the ability of MGA-Net to comprehend projective relations better, particularly when the target object is not apparent. An example of Proximity relations is in Figure~\ref{2387165.jpg} where LLaVA and Grounding DINO return the shore that is ‘near’ the murky water, but MGA-Net fails to do so.
\subsection{Multiple Spatial Relations}
For ‘Two-and' category expressions, the models sometimes only satisfy one of the spatial clauses. This often happens if multiple objects of the same class are in the image. For example, in Figure~\ref{1159745.jpg}, the output baseball player is to the left of the black helmet but is not to the right of the home plate.

Similarly, for 'Two-chained' category expressions, the models sometimes do not consider the entire expression. For example, in Figure~\ref{2324708.jpg}, MGA-Net and LLaVA return the ‘log that is behind the large bear’, and Grounding DINO returns the bear itself. None of the models consider the ‘large branch’ part of the expression, which should have been the output.

Finally, for 'Two-or' category expressions, the model might pay attention to only one spatial clause. Consequently, it returns an object satisfying that clause but not the additional attributes mentioned in the expression. For example, in Figure~\ref{285860.jpg}, the model returns the monitor, which is to the ‘left of the keyboard’, but it does not satisfy the color attribute.
\subsection{Negation}
Figures \ref{2363953.jpg} and \ref{2345193.jpg} show two cases where all models fail to recognize negation. In \ref{2363953.jpg}, we can observe that while MGA-Net is wrong, LLaVA is close to the ground truth but partially covers the target object (high precision, low recall). In \ref{2345193.jpg}, while LLaVA is wrong, MGA-Net is closest to the ground truth but covers an excess area (low precision, high recall).
\section{Conclusion}
Spatial reasoning and understanding is an area in which the latest VLMs have shown signs of struggle. We evaluate the spatial understanding of a variety of models using the referring expression comprehension task because it requires explicit grounding of complex linguistic expressions in the visual modality. We picked multiple models including Vision-Language models (LLaVA, Grounding DINO) as well as task-specific models (MGA-Net). We observed that the VLMs that are trained in the wild with visual and textual data perform worse in grounding. All models have challenges in Directional relations. However, the VLMs do better in vague relations such as proximity while the task-specific models are better in geometrically well-defined relations such as left and right. While using spatial relations helps in grounding, using multiple relations makes the reasoning more challenging for all models, with a higher impact on VLMs. MGA-Net handles complex spatial expressions better than VLMs due to its compositional learning architecture. In the presence of visual complexity, all models face challenges but LLaVA struggles the most due to lack of grounded pre-training. Finally, both VLMs and task-specific models struggle with grounding expressions that include negation. These findings shed light on the gaps for future work on Vision-language models.

\section{Future Directions}
We observed that MGA-Net handles expressions with varying spatial complexity better than the VLMs due to its soft attention module which decomposes the expression into its semantic components for compositional reasoning. This highlights the decomposition of complex spatial expressions as a potential path forward to help VLMs generalization.~\cite{compsurvey} discuss multi-modal transformer models introduced by~\cite{sikarwar2022can},~\cite{qiu2021systematic} and techniques such as weight sharing across transformer layers or ‘Pushdown layers’ with recursive language understanding~\cite{murty2023pushdown} as an alternative to self-attention to aid compositional reasoning. Another promising direction is Neuro-symbolic processing~\cite{kamali2024nesycoco,hsu2024}, which involves generating symbolic programs from expressions using LLMs and conducting explicit symbolic compositions before grounding into visual modality. We plan to explore integrating such techniques with VLMs to improve their spatial compositional reasoning capabilities.

Another issue to address is the VLMs' inability to comprehend negations. MGA-Net's improved performance over Grounding DINO due to the presence of negated expressions in the training data motivates us to explore the augmentation of training/instruction tuning data of VLMs with synthetically generated negated expressions. Additionally, we also plan to formulate contrastive learning objectives to penalize the model when it fails to comprehend negations.

\bibliographystyle{named}
\bibliography{arxiv}

\appendix

\section{Appendix}

\subsection{Description of spatial categories}\label{app: cat ex} 
For our analysis, we utilize the spatial categories introduced by ~\cite{marchi} and replace the 'Cardinal Direction' category with 'Absolute'. The descriptions and examples for the chosen categories are as follows:
\begin{enumerate}
    \item \textbf{Absolute}: Consists of relations that describe the location of an object in an absolute manner and not in relation to another object. \\
    Eg:- man on the right that is standing and wearing gray pant 
    \item \textbf{Adjacency}: Consists of relations that describe the physical proximity of two objects. \\
    Eg:- The large poster that is leaning against the wall
    \item \textbf{Directional}: Consists of dynamic action verbs / directional relations. The interpretation of these relations heavily relies on the configuration of the involved objects and/or the dynamic spatial relationship between them. \\
    Eg:- The gray car that is driving down the road
    \item \textbf{Orientation}: Consists of relations which describe the orientation of an object w.r.t another object. \\
    Eg:- The sitting dog that is facing the window that is to the right of the mirror
    \item \textbf{Projective}: Consists of relations that indicate the concrete spatial relationship between two objects, i.e., these relations can be quantified in terms of the coordinates of the two objects. \\
    Eg:- The black oven that is above the drawer
    \item \textbf{Proximity}: Consists of relations that indicate that two objects are near each other without giving a specific directional relationship. \\
    Eg:- The blue chair that is close to the white monitor
    \item \textbf{Topological}: Consists of relations that indicate the broader arrangement or the containment of an object w.r.t another object \\
    Eg:- The silver train that is at the colorful station
    \item \textbf{Unallocated}: Consists of relations that cannot be allocated to any of the above categories.
\end{enumerate}

\subsection{Other Models}\label{app: other models}
In our analysis, we also experimented with InstructBLIP \cite{instructblip} and OpenFlamingo \cite{openflamingo} models. These models are general-purpose VLMs with InstructBLIP working in the zero-shot model and OpenFlamingo in the few-shot mode. However, neither of the models could provide meaningful outputs for the task. In this section, we disclose the prompts that we used for these two models and the outputs obtained for the prompts:

\subsubsection{InstructBLIP}
For InstructBLIP, we designed three prompts for the REC task. They are as follows:
\begin{enumerate}
    \item Bounding Boxes: {bounding box list}; Referring Expression: {Refexp}; The index of the output bounding box is:
    \item Bounding Boxes: {bounding box list}; Referring Expression: {Refexp}; The coordinates of the output bounding box are:
    \item Provide the bounding box coordinates for: "{Refexp}"
\end{enumerate}
In both prompts, the ‘bounding box list’ consists of the coordinates of the detected bounding boxes in the image being passed as the input, along with indices for each bounding box, starting from ‘1’. But for the third prompt, the model has no access to pre-detected candidate bounding boxes in the image. While the expected output for the first prompt was the index of the correct bounding box, for the other 2 prompts it was the bounding box coordinates as the output.

The bounding box format is [x1, y1, x2, y2], where (x1, y1) is the bottom left corner and (x2, y2) is the top right corner of the box. The coordinate values are a fraction of the total length/width of the image according to the position of the coordinate.

Unfortunately, none of the three prompts gave a satisfactory output. The outputs were as follows: \\
\textbf{Prompt 1}: The outputs were mostly wrong. Sometimes, the model also gave ‘0’ as the output, even though it is not a valid index. \\
\textbf{Prompt 2}: The output did not return meaningful coordinates in most cases. But in the few instances that it did, they were mostly wrong. Example outputs when the model could not return meaningful coordinates are: 
\begin{itemize}
\item \{‘1’: [0.16, 0.55], ‘2’: [0.32, 0.47], ‘3’: [0.55, 0.6], ‘4’: [0.21, 0.06],...
\item 0.9, 0.53, 0.93, 0.57, 0.0, 0.39,...
\end{itemize}
\textbf{Prompt 3}: The model could not understand the task and it just paraphrased parts of the prompt instead of giving the coordinates as the output. Example prompts and outputs are:
\begin{itemize}
\item \textbf{Prompt:} Provide the bounding box coordinates for: "The large poster that is leaning against the wall" \\
\textbf{Output}: what is the bounding box coordinates for the large poster that is leaning against the wall
\item \textbf{Prompt:} Provide the bounding box coordinates for: "The young man that is leaning against the wall" \\
\textbf{Output}: is standing in an elevator. the young man that is leaning against the wall is standing in an elevator
\end{itemize}

\subsubsection{OpenFlamingo}
As discussed before, OpenFlamingo is a few-shot prompting model. We tested the model in 2 and 3-shot settings. \\\\
\textbf{Prompt 1}:
\begin{itemize}
    \item \textbf{Example output format:} \textless image\textgreater Bounding Boxes:{bounding box list}; Expression: {Refexp}; Correct Bounding Box:"ID"\textless\textbar endofchunk\textbar\textgreater
    \item \textbf{Query format:} \textless image\textgreater Bounding Boxes:{bounding box list}; Expression: {Refexp}; Correct Bounding Box:"
\end{itemize}
‘bounding box list’ took the list of candidate bounding boxes in the image as input, in the same format as for InstructBLIP (discussed in the previous section). The expected output was the index of the correct bounding box. However, we observed that irrespective of the query, the model gave the same output index for the same set of prompting examples. \\\\
\textbf{Prompt 2}: 
\begin{itemize}
    \item \textbf{Example output format:} \textless image\textgreater Expression: {Refexp}; Correct Bounding Box:[Bounding box coordinates]\textless\textbar endofchunk\textbar\textgreater
    \item \textbf{Query format:} \textless image\textgreater Expression: {Refexp}; Correct Bounding Box:[
\end{itemize}
‘bounding box list’ takes the same input as explained for Prompt 1. But instead of expecting the index, we expect the coordinates of the bounding box as the output. The format of the bounding box is the same as explained for InstructBLIP in the previous section. However, the model failed to give meaningful coordinates as output in most cases, similar to InstructBLIP. When it did give meaningful coordinates, the outputs were mostly wrong.

\subsection{Related Works Data}
Table~\ref{table:prev lit} provides the evaluation task, list of evaluated models, the benchmark used, and the properties of the benchmarks used in the works that include spatial relation analysis. Table~\ref{table:prev spat} provides additional information such as the properties of the images and the spatial complexity of the benchmarks used in these works. Towards the end of both tables, we also include the said characteristics of our work for comparison. Among the mentioned works, only our work focuses on category-wise spatial analysis while using complex spatial expressions.

\begin{table*}[!p]
\centering
\begin{tabular}{p{1.9cm}p{2cm}p{2.4cm}p{3.6cm}p{3.7cm}}
\toprule
Work & Bench mark & Evaluation Tasks & Models & Text Source \\
\midrule
  Rows for data entries
~\cite{pic1k} & New: SpatialEval & VQA & LLaVA, InstructBLIP, GPT-4V and 4o, Bunny, etc. & Human and Template Annotation \\
~\cite{blink} & New: BLINK & Image-Caption agreement & MiniGPT-4, OpenFlamingo, InstructBLIP, LLaVA, GPT-4V, etc. & Human annotation with constraints \\
~\cite{cb300k} & New: CB-300K & Multi-round multimodal grounding & LLaVA, InstructBLIP, VisionLLM, Kosmos-2, GPT4RoI, etc. & GPT-4-generated questions and answers for the conversation \\
    ~\cite{llava} & New: LLaVA-Bench & VQA & LLaVA, BLIP, OpenFlamingo & Questions generated by GPT-4 / ChatGPT along with human-curated descriptions \\
~\cite{vsr} & New: VSR & Image-caption agreement & Visual BERT, CLIP, LXMERT, ViLT & Human-annotation with constraints \\
~\cite{mmbench} & New: MMBench & Multi-Choice VQA & OpenFlamingo, LLaVA, MiniGPT4, InstructBLIP, Qwen-VL, etc. & Human-annotated; post-processing by evaluating with LLMs and VLMs \\
~\cite{seedbench} & New: SEEDBench & Multi-Choice VQA & OpenFlamingo, BLIP-2, LLaVA, InstructBLIP, MultimodalGPT, etc. & Generated by ChatGPT / GPT-4; post-processing by Humans, LLMs \\
~\cite{mmvet} & New: MM-Vet & VQA & OpenFlamingo, LLaVA, MiniGPT-4, InstructBLIP, GPT-4V, etc. & Manually annotated \\
~\cite{roschpi} & Datasets: MSCOCO and Visual Genome & Image-caption agreement & LXMERT & Human annotation \\
~\cite{whatsup} & New: What'sUp & Multi-choice Image Captioning & CLIP, RoBERTaCLIP, XLVM, BLIP, FLAVA, etc. & Template annotation \\
~\cite{sr2d} & New: SR2D & Text-to-image generation & GLIDE, DALLE-mini and v2, CogView2, Stable Diffusion, etc. & Template annotation \\
~\cite{clipcdsm} & Synthetic Dataset & Multi-choice Image Captioning & CLIP, CSP, CDSMs & Human and template annotation \\
~\cite{subramaniamreclip} & Synthetic Dataset & Multi-choice Image Captioning & CLIP RN50 versions, CLIP ViT-B versions, ALBERT, etc. & Human-annotation with constraints \\
~\cite{film} & Synthetic Datasets & Image-caption agreement & FiLM, CNN-LSTM, CNN-LSTM SA & Template annotation \\
Ours & Dataset-CopsRef & REC & MGA-Net, Grounding DINO, LLaVA & Template annotation \\
\bottomrule
\end{tabular}
\caption{Statistics - Previous literature on evaluating VLMs for spatial reasoning and understanding. In the ‘Bench mark' column, New: benchmarks introduced by the authors of those works. Synthetic Datasets: datasets generated by the authors for their analysis but not released as benchmarks.
}
\label{table:prev lit}
\end{table*}

\begin{table*}[!htb]
\centering
\begin{tabular}{p{2cm}p{2cm}p{3cm}p{2.5cm}p{1.3cm}p{1.1cm}p{1.1cm}}
\toprule
 Work & Image Type & Image Source & Spatial & No. of & Spatial & Deeper \\
 & & & Complexity & Spatial & Relation & Spatial \\
 & & & & Relations & Focus & Analysis \\
\midrule
~\cite{pic1k} & Real-world & SA-1B Dataset (~\cite{sa-1b}) & Questions - Simple, Text - Complex & - & Yes & No \\
~\cite{blink} & Real-world & MSCOCO Dataset & Captions - Simple & - & No & No \\
~\cite{cb300k} & Real-world & Visual Genome & Questions / Answers - Complex & 80 & No & No \\
~\cite{llava} & Real-world & 24 diverse images & Questions / Answers - Simple & 37 & No & No \\
~\cite{vsr} & Real-world & MSCOCO Dataset & Captions - Simple & 66 & Yes & Yes \\
~\cite{mmbench} & Real-world & Internet sources & Questions / Options - Medium & - & No & No \\
~\cite{seedbench} & Real-world & CC3M Dataset & Questions / Options - Simple & 86 & No & No \\
~\cite{mmvet} & Real-world & Online Sources, VCR & Questions / Answers - Medium & 15 & No & No \\
~\cite{roschpi} & Real-world & MSCOCO and Visual Genome Datasets & Captions - Medium & 28 & Yes & No \\
~\cite{whatsup} & Real-world & Objects captured in a controlled environment & Captions - Simple & 6 & Yes & Yes \\
~\cite{sr2d} & Real-world & MSCOCO Dataset & Captions - Simple & 4 & Yes & Yes \\
~\cite{clipcdsm} & Synthetic & Generated using ‘CLEVR’ process & Captions - Simple & 4 & No & Yes \\
~\cite{subramaniamreclip} & Synthetic & Generated using ‘CLEVR’ process & Captions - Simple & 4 & Yes & No \\
~\cite{film} & Synthetic & ShapeWorld dataset & Captions - Simple & 8 + their superlatives & No & Yes \\
Ours & Real-world & CopsRef Dataset & Referring expressions - Complex & 51 & Yes & Yes \\
\bottomrule
\end{tabular}
\caption{Spatial analysis - Previous literature on evaluating VLMs for spatial reasoning and understanding. In the Spatial Complexity column, simple: text with a maximum of 2 objects and 1 spatial relation, medium: a maximum of 3 objects and 2 spatial relations, complex: more than 3 objects and 2 spatial relations. For the 'No. of spatial relations' column, the relation types are taken from all the works mentioned in this table and~\protect\cite{marchi}, in addition to the relations in Table 2 of the main paper.}
\label{table:prev spat}
\end{table*}

\end{document}